\documentclass[a4paper, 10pt, conference]{svproc}  
\usepackage{times}
\usepackage{float}
\usepackage{wrapfig}

\usepackage{amsmath}

\usepackage{soul}
\usepackage{natbib}
\usepackage{graphicx}
\usepackage{comment}
\usepackage[font=small]{caption}
\usepackage{subcaption}
\newtheorem{prop}{Proposition}
\usepackage{booktabs}
\usepackage{tabularx}
\usepackage{hyperref}
\newcolumntype{Y}{>{\centering\arraybackslash}X}
\newcolumntype{Z}{>{\centering\hsize=.8\hsize}X}

\usepackage[dvipsnames]{xcolor}

\title{ 
Diagnostic Runtime Monitoring with Martingales 
}

\titlerunning{Runtime Monitoring}

\author{Ali Hindy\inst{1} \and Rachel Luo\inst{1} \and 
Somrita Banerjee\inst{1} \and Jonathan Kuck\inst{2} \and Edward Schmerling\inst{1} \and Marco Pavone\inst{1} 
\thanks{The NASA University Leadership Initiative (grant \#80NSSC20M0163) provided funds to assist the authors with their research, but this article solely reflects the opinions and conclusions of its authors and not any NASA entity.}
}

\authorrunning{Ali Hindy et al.} 

\tocauthor{Ali Hindy, Rachel Luo, Somrita Banerjee, Jonathan Kuck,
Edward Schmerling, and Marco Pavone}

\institute{Stanford University, Stanford, CA, USA,\\
\email{\{ahindy, rsluo, somrita, schmrlng, pavone\}@stanford.edu},
\and
Dexterity AI, Redwood City, CA, USA,
\email{jonathan@dexterity.ai}}

\begin{document}

\maketitle
\thispagestyle{empty}
\pagestyle{empty}

\begin{abstract}
Machine learning systems deployed in safety-critical robotics settings must be robust to distribution shifts. However, system designers must understand the \textit{cause} of a distribution shift in order to implement the appropriate intervention or mitigation strategy and prevent system failure. In this paper, we present a novel framework for diagnosing distribution shifts in a streaming fashion by deploying multiple stochastic martingales simultaneously. We show that knowledge of the underlying cause of a distribution shift can lead to proper interventions over the lifecycle of a deployed system. Our experimental framework can easily be adapted to different types of distribution shifts, models, and datasets. We find that our method outperforms existing work on diagnosing distribution shifts in terms of speed, accuracy, and flexibility, and validate the efficiency of our model in both simulated and live hardware settings. 

\end{abstract}


\section{Introduction}

Modern learning-enabled systems deployed in the real world are susceptible to failure if they encounter test-time inputs that do not follow the same distribution as the training-time inputs. However, the complexity of these learned systems leads to significant challenges in diagnosing problems that arise during operation \citep{nandi1999fault}. Runtime monitors can alert users when issues arise, but pinpointing the underlying cause of the issue is difficult. For instance, a problem could stem from a suboptimal training process, during which the model was not exposed to a sufficiently diverse or representative dataset, leading to poor generalization capabilities. Alternatively, the problem might originate from the operational environment, where real-time conditions differ from those encountered during training (e.g. due to sensor degradation or environmental changes).

\begin{figure}
    \centering
    \includegraphics[clip, width=0.5\textwidth]{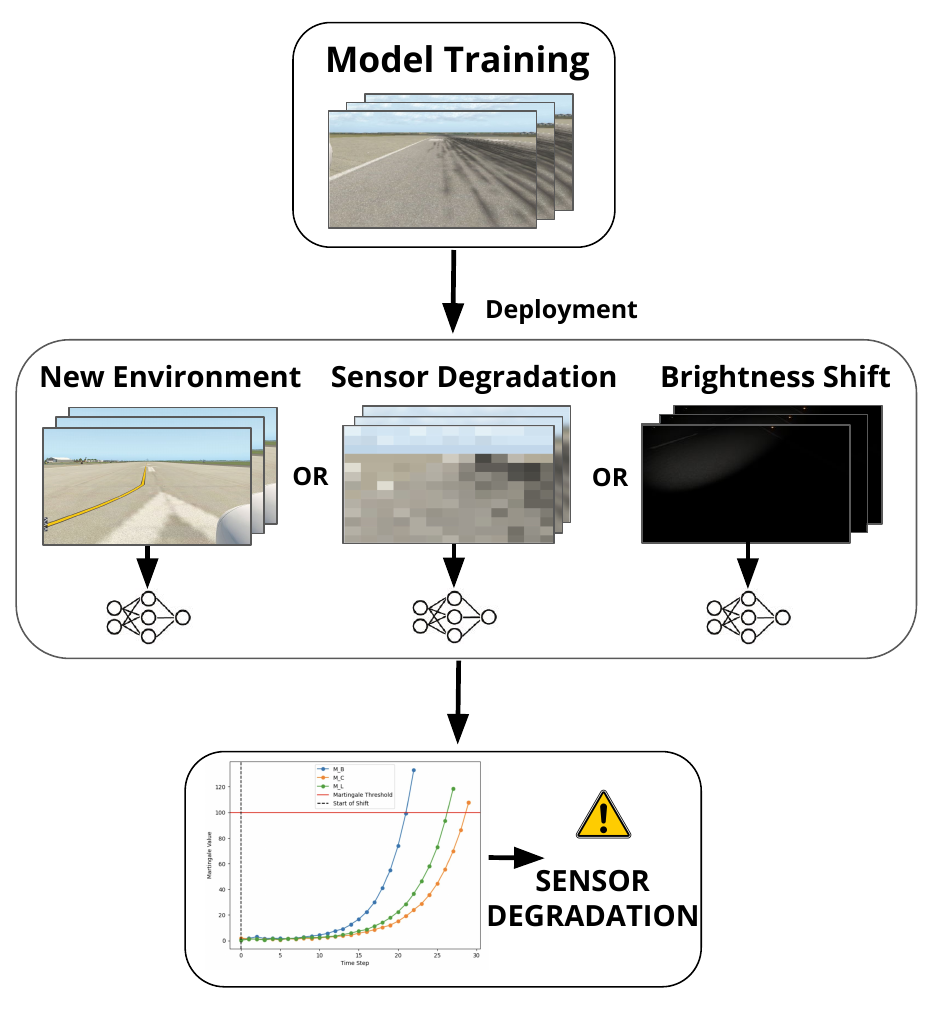}
    \caption{Overview of our high-level approach.  Learning-enabled robotics systems are trained on data from a finite set of environments. When deployed, these systems may operate in distribution-shifted conditions, resulting in erroneous predictions. Our method issues an alert if conditions change, and alerts users of a probable underlying cause. Knowledge of the underlying cause informs the choice of the proper intervention method to restore system performance.}
    \label{fig:main_figure}
\end{figure}

As an example, consider a camera-driven robot such as an autonomous aircraft using a PID controller to taxi along the centerline of a runway, where learned models estimate the cross-track and heading errors relative to that centerline. In this setting, problems that could cause the robot to fail include new situations that the robot has not previously encountered (e.g. the autonomous aircraft taking off from a runway at a new airport), or sensor degradation. In the case of new situations that the robot has not previously encountered, there is an offline problem (i.e. the unrepresentative training data), and we expect the distribution of both the system inputs and system outputs to change. For example, a new airport will look different than the original airport, so the input images will change; the width of the new runway will be different, so the distribution of deviations from the centerline will also change. In the case of sensor degradation, there is an online operational problem (the system that's running has changed), and we expect the distribution of system inputs to change, but not the distribution of system outputs. The input images may look grainier, but the distribution of the cross-track and heading errors will remain the same.  

In the example described above, using two runtime monitors that issue alerts in case of distribution shift --- one over the system inputs and one over the system outputs --- could differentiate between the two types of problems. More generally, if system designers identify that certain features correspond to specific issues, they can strategically place targeted runtime monitors over those features. This approach facilitates a more efficient troubleshooting process, since the activated monitors provide useful insights into which aspects of the system are contributing to the issue. 
By quickly detecting the cause of an issue, an appropriate intervention can then be applied. For example, in the case of a new runway at a new airport, the appropriate intervention is to collect additional data and perform weighted retraining. In the case of sensor degradation, the appropriate intervention is to replace the sensor. 

In this work, we present a framework for determining the cause of a distribution shift, using multiple martingale-based runtime monitors based on the method described in \cite{luo2023online}. As in~\citep{luo2023online}, we focus on episodic situations (e.g. for a plane repeatedly taxiing down a runway during a continuous deployment, each taxiing sequence can be considered an episode drawn from a task distribution). Our method can be applied to any online learning setting, such as deploying autonomous vehicles, recognizing spoken language, or evaluating the performance of large language models. An overview of our system is shown in Figure~\ref{fig:main_figure}.  

The contributions of our work are as follows: 1) We introduce a method that quickly alerts users when a distribution shift has occurred and enables rapid diagnosis of the underlying problem. 2) Our approach detects the cause of a distribution shift faster than prior work, and leads to better outcomes over the lifecycle of a robot. 3) We empirically evaluate our approach on photorealistic simulations of an autonomous aircraft taxiing down a runway with a camera perception module and in hardware on a free-flyer space robotics testbed for vision-based navigation. In these experiments, our method detects the cause of a distribution shift up to five times faster than prior work, and our method is effective at eliminating system failures due to causal misidentification experienced by a baseline system.

\section{Background and Related Work}
\label{sec:related}
\subsection{Distribution Shift Detection}

\paragraph{Traditional Approaches. }
The challenge of detecting distribution shift has been explored by both the machine learning and statistics communities. Traditional approaches typically rely on statistical hypothesis testing to assess whether the test-time distribution differs from the training distribution. Most of these methods focus on covariate shift, where the distribution of input data changes while the conditional distribution of labels given inputs remains constant ~\citep{GrettonBorgwardtEtAl2012, Rabanser2019FailingLA, Kulinski2020FeatureSD, Kamulete2021TestFN, chang2021mitigating}. Some methods have also been proposed for label shift, where the distribution of labels changes but the conditional distribution of inputs given labels remains the same. For example,  \citep{alexandari2020maximum} propose a maximum likelihood algorithm for detecting and correcting label shift; \citep{Rabanser2019FailingLA} use an efficient weight estimator to provide a generalization bound for the label shift problem; \citep{lipton2018detecting} introduce black box predictors and a score-based test statistic for detecting label shift. However, these methods are generally designed for offline (batch) scenarios, and applying them online in a robotics setting may either compromise their statistical guarantee or lead to statistical inefficiency. Furthermore, these methods focus on detecting only a single type of distribution shift.

\paragraph{Martingale-Based Approaches. }
A martingale (Definition~\ref{def:martingale}) is a stochastic process (a sequence of random variables) where the conditional expectation of the next value, given all previous values, is the same as the most recent value \citep{books/daglib/0073491}. Doob's Inequality (Proposition~\ref{prop:doob}), states that the probability that a martingale grows very large is very small \citep{books/daglib/0073491}.

\begin{definition}[Martingale]
\label{def:martingale} 
A martingale is a sequence of random variables $M_1$, $M_2$, \dots, such that $E[|M_{n}|] < \infty$ and
$ E[M_{n+1} | M_1, \dots, M_n] = M_n $ for all $n$.
\end{definition}

\begin{prop}[Doob's Inequality]
\label{prop:doob}
For a martingale $M_n$ indexed by an interval $[0, N]$, and for any positive real number $C$, it holds that
\begin{equation*}
    \mathrm{Pr}\left[\sup_{0 \leq n \leq N} M_n \geq C\right] \leq \frac{E[\max(M_N,0)]}{C}.
\end{equation*}
\end{prop}

An approach for detecting distribution shift in online settings, introduced by \citet{vovk2020testing}, uses conformal martingales to test for exchangeability. 
Other works following this approach include \citep{eliades2020histogram, volkhonskiy2017inductive, fedorova2012plug, ho2005martingale, podkopaev2021tracking, hu2020distribution}. These methods apply conformal prediction to obtain p-values for each test sample, which are then used to construct a martingale. When the martingale exhibits significant growth, it indicates a probable distribution shift. \citep{vovk2020testing, Vovk2021TestingRO, vovk2021conformal, vovk2021retrain} introduce different p-value calculations for different types of distribution shifts. While these works demonstrate good efficiency on lower dimensional data (e.g. on the USPS dataset~\citep{uspsdataset}, which contains 11-dimensional feature vectors), they often struggle in more complex or higher-dimensional robotics settings, such as those involving image data.

More recently, \citet{luo2023online} introduce a learned, martingale-based runtime monitor that detects distribution shifts quickly for high-dimensional data, with guarantees limiting the number of false alarms. This method trains a neural network model to distinguish between older and more recent samples, and issues a warning when the model is consistently able to predict recency. A key component of this method is a martingale-based test statistic designed to grow if the distribution shifts. 
However, this work by default applies a martingale monitor only over the system inputs, making it impossible to distinguish between different underlying problems. In our work, we introduce a framework using multiple learned, martingale-based runtime monitors, and apply them over different feature spaces to improve problem diagnosis.

\subsection{System Fault Diagnosis}
Many works leverage deep learning for system fault detection, using large models to detect one type of system fault \cite{wu2008expert, hajnayeb2008design, merainani2018novel, zabihi2019fault, zheng2019sparse}. However, these papers deploy large models with long training times that are impractical for live robotics systems, and they focus on diagnosing faults in specific mechanical systems in offline settings that are not easily generalizable. Our method, meanwhile, works with lightweight models suitable for online robotics settings. 

Other learned methods using simpler models for system fault detection include \cite{toma2020bearing, 9467525, lei2020applications, hajnayeb2008design, yang2004art, li2016new}.
However, these methods generally deal with low-dimensional data and require access to the true labels. Our method, meanwhile, works with high-dimensional data without access to true labels.

Some methods for fault diagnosis in online settings include \citep{xingxin2022research, seera2013online, netti2020machine, kang2018machine}; however, these works focus on anomaly detection (i.e. experiencing a single rare event), while our work focuses on gradual distributional shifts.

\section{Identifying Types of Distribution Shift} 
\label{sec:method}

We present a framework for quickly identifying distribution shifts in episodic robotics settings that provides actionable information for targeted interventions.  Our method builds on the work of \citet{luo2023online} and also leverages stochastic martingale-based runtime monitors.  Our critical contribution is to use \textit{multiple} martingale-based monitors, deployed simultaneously with each monitoring a different type of distribution shift.  This enables our system to detect distribution shifts more quickly, and moreover the monitors each probe a distinct intermediate result of the robotics system.  This enables system designers to identify system components that are behaving abnormally before failure of the entire robotics system, develop appropriate interventions, and differentiate between types of distribution shift.

\subsection{Problem Setup }
Consider an autonomous system that interacts with the environment.  The system's behavior is defined by function $f: \mathcal{X} \rightarrow \mathcal{Y}$ which maps sensor inputs $x^t \in \mathcal{X} $ at time $t$ to action $y^t \in \mathcal{Y}$.  This function can be expressed as the composition of $K$ functions, $f = f_K \circ \dots \circ f_2 \circ f_1$, with $K-1$ intermediate results $I_1, I_2, ... I_{K-1}$, where $I_1 = f_1(x), 
I_k = f_k(I_{k-1}) \text{ for } 1 < k < K, \\
y = f_K(I_{K-1})$.

\begin{figure*}[!tb]
    \centering
    \begin{subfigure}[c]{0.325\textwidth}
        \centering
        \includegraphics[width=\textwidth]{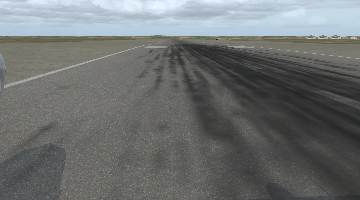}\vspace{1mm}
        \caption{Nominal image}
        \label{fig:camera_correct}
    \end{subfigure}
    \begin{subfigure}[c]{0.325\textwidth}
        \centering
        \includegraphics[width=\textwidth, height=2.2cm]{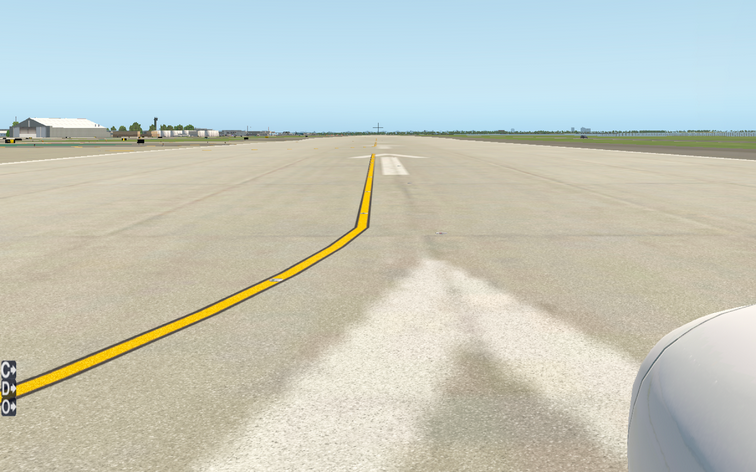}\vspace{1mm}
        \caption{New Airport}
        \label{fig:camera_perturbed}
    \end{subfigure}
    \begin{subfigure}[c]{0.325\textwidth}
        \centering
        \includegraphics[width=\textwidth]{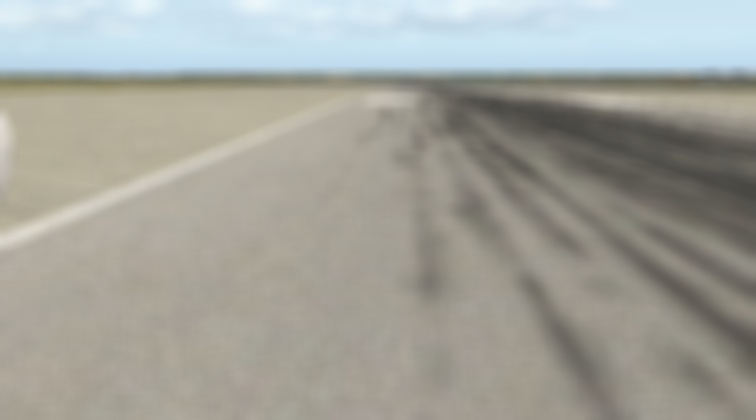}\vspace{1mm}
        \caption{Sensor Degradation}
        \label{fig:xplane_camera_results}
    \end{subfigure}
    \caption{Images generated from the X-Plane 11 flight simulator, with (\ref{fig:camera_correct}) a standard camera angle,  (\ref{fig:camera_perturbed}) a shifted environment, and (\ref{fig:xplane_camera_results}) sensor degradation.  We define separate martingales to identify each type of shift.}
    \label{fig:xplane_shifts}
\end{figure*}

In our problem setup we are given a dataset of $N$ historical inputs sampled from a single distribution, $D_{\mathrm{orig}} = (X_1, X_2, \cdots, X_N)$.  Each $X_i$ represents a sequence of sensor inputs recorded during an episode of the robot interacting with the environment, e.g. $X_i = (x_i^1, x_i^2, \dots, x_i^{l_i})$.  During the robot's deployment, in a potentially novel environment, we observe new sequences of sensor inputs $D_{\mathrm{new}} = (X_{N+1}, X_{N+2}, \dots)$.  Again each $X_i$ in $D_{\mathrm{new}}$ represents a sequence of inputs recorded during an episode of the robot interacting with the environment.  

As shown in \citet{luo2023online} it is possible to design a series of test functions over system inputs

\begin{align*}
& \psi^1_j: D_{\mathrm{orig}}, X_{N+1}, \cdots, X_{j} \mapsto \lbrace T, F \rbrace  \\
& \forall j = N + 1, N + 2, \ldots,
\end{align*}
where the output $T$(rue) indicates that we have identified a distribution shift over the system inputs, while $F$(alse) indicates that we have failed to identify a distribution shift over the system inputs.

In this work we additionally aim to design test functions over the $K-1$ intermediate results and the final action.  These test functions will issue an output alert when a distribution shift is identified over the intermediate results or the final action.  The test functions will take the form
\begin{align*}
& \psi^k_j: D_{\mathrm{orig}}, f^{1 \rightarrow k}_j (X_{N+1}), \cdots, f^{1 \rightarrow k}_j (X_{j}) \mapsto \lbrace T, F \rbrace  \\
& \forall j = N + 1, N + 2, \ldots,
\end{align*}
where $1 \leq k < K$ for test functions over the intermediate results and $k=K$ for the test function over the output actions.  We define $f^{1 \rightarrow k}_j : \mathcal{X}^{l_j} \rightarrow \mathcal{Y}^{l_j}$ as \footnote{That is, the domain of $f^{1 \rightarrow k}_j$ is the $l_j$-fold Cartesian product of $\mathcal{X}$ and the codomain is the $l_j$-fold Cartesian product of $\mathcal{Y}$, where $l_j$ is the length of episode $j$.}
\begin{equation*}
f^{1 \rightarrow k}_j(X_{j}) := \left( 
\begin{aligned}
&f_k \circ \dots \circ f_2 \circ f_1 (x^1_{j}), \\
&f_k \circ \dots \circ f_2 \circ f_1 (x^2_{j}), \\
&\vdots \\
&f_k \circ \dots \circ f_2 \circ f_1 (x^{l_j}_{j})
\end{aligned}
\right).
\end{equation*}

Additionally, each test function is designed to be $\epsilon$-sound (\citet{luo2023online}), meaning that if the historical data $D_{\mathrm{orig}}$ and the new data $D_{\mathrm{new}}$ are drawn from the same distribution, then with high probability (1 - $\epsilon$) the test will not issue a warning.  Formally, when $D_{\mathrm{new}}$ and $D_{\mathrm{orig}}$ are exchangeable, then 
\begin{align*}
    \Pr\big[\exists j, \psi^k_j (D_{\mathrm{orig}}, f^{1 \rightarrow k}_j (X_{N+1}), \cdots, f^{1 \rightarrow k}_j (X_{j}) ) = T \big] \leq \epsilon.
\end{align*}

\subsection{Proposed Framework }
The key intuition behind martingale-based monitors is that it is impossible for any predictor to reliably distinguish between two samples drawn from the same distribution.  Consider drawing two samples, one from $D_{\mathrm{new}}$ and the other from from $D_{\mathrm{orig}}$.  If there has been no distribution shift (so these distributions are in fact the same), then no predictor can do better than random chance (e.g. a Bernoulli variable with parameter $p = 0.5$).

We adapt the martingale-based monitor defined in \citet{luo2023online} to design the test functions $\psi^1_j$ over system inputs.  This monitor relies on a neural network $f^1_{NN} : \mathcal{X} \times \mathcal{X} \rightarrow \{0, 1\}$ that is trained to predict whether two inputs come from the same distribution.  Additionally we design test functions $\psi^k_j$ for $k = 1, \dots, K$ over all $K-1$ intermediate results $I_1, I_2, ... I_{K-1}$ and over the final output action.  For the test functions over intermediate results, the neural network is modified to take as input two intermediate results, $f^k_{NN} : I_k \times I_k \rightarrow \{0, 1\}$.  
For the monitor over final output actions, the neural network is modified to take as input two actions, $f^K_{NN} : \mathcal{Y} \times \mathcal{Y} \rightarrow \{0, 1\}$.  Note that our method is agnostic to the specific choice of model architecture (but for details about the model architectures that we used in our experiments, refer to the Appendix.

To design the test functions, let us first define the indicator variables
\begin{equation*}
\label{eq:indicator}
    Z^k(j) = 
    \begin{cases}
        1\hspace{0.5cm} \text{if } f^k_{NN} \text{ predicts correctly for test example } j \\
        0\hspace{0.5cm} \text{otherwise}.
    \end{cases}
\end{equation*}
Then let us define the martingales
\begin{equation*}
\label{eq:martingale}
    M^k_n  = \frac{(e^{t \cdot S^k_n})}{((q + p e^t)^n)},
\end{equation*}
where $S^k_n = \sum_{j=1}^{n} Z^k(j)$, $p = q = 0.5$, and we use $t = 1$ (Equation 4 in \citet{luo2023online}).  Using these martingales we define the test functions as
\begin{equation}
\label{eq:psi-def}
   \psi^k_j( D_{\mathrm{orig}}, X_{N+1}, \cdots, X_{j}) = 
    \begin{cases}
        T\hspace{0.5cm} \text{if } M^k_n  \geq C \\
        F\hspace{0.5cm} \text{otherwise}.
    \end{cases}
\end{equation}
Using a threshold of $C = 100$ we are guaranteed a false positive rate of $\le 0.01$ for each test function because (Lemma 2 in \citet{luo2023online})
\begin{equation*}
    \mathrm{Pr}\left[\sup_{0 \leq n \leq N} M^k_n  \geq C\right] \leq \frac{1}{C}.
\end{equation*}

This allows our framework to propose a mitigation strategy tailored to the specific failure mode identified by the martingale monitor that signaled the distribution shift. 

In order to identify a particular type of distribution shift, we need to construct different martingales, each with representative inputs for our test functions. As an illustrative example, we consider distribution shifts that may occur for images taken by an external camera, as shown in Figure{~\ref{fig:xplane_shifts}}. During deployment, we may encounter sensor degradation, which is a shift that results in a change to the input images, so we feed in representative input features to the classifier. If a distribution shift is detected using these input features, we know that the distribution of our input images has changed, and a good mitigation strategy would be to replace the sensors. Similarly, if we encounter a new environment, such as a new airport, our output labels will change, as the plane taxis with a new trajectory. Therefore, we feed in output features to the classifier, and if we detect a distribution shift using the output features, we know that the distribution of our output labels has changed. Here, a good mitigation strategy would be to gather additional data and perform a weighted retraining. The key insight is to use knowledge of a likely type of distribution to inform the relevant chosen features.

\section{Experiments }
\label{sec:experiments}
  
We empirically evaluate our method on photorealistic simulations of an autonomous aircraft taxiing down a runway using camera-based perception and in hardware on a free-flyer space robotics testbed for vision-based navigation. We show that our method detects distribution shifts and identifies the type of distribution shift more quickly than existing methods. We also show that the knowledge of distribution type provided by our method allows us to use targeted interventions, which leads to better results than generic interventions, and better outcomes over the robot lifecycle compared to a scheduled maintenance approach. 
\begin{figure}
    \centering
    \includegraphics[width=0.45\textwidth]{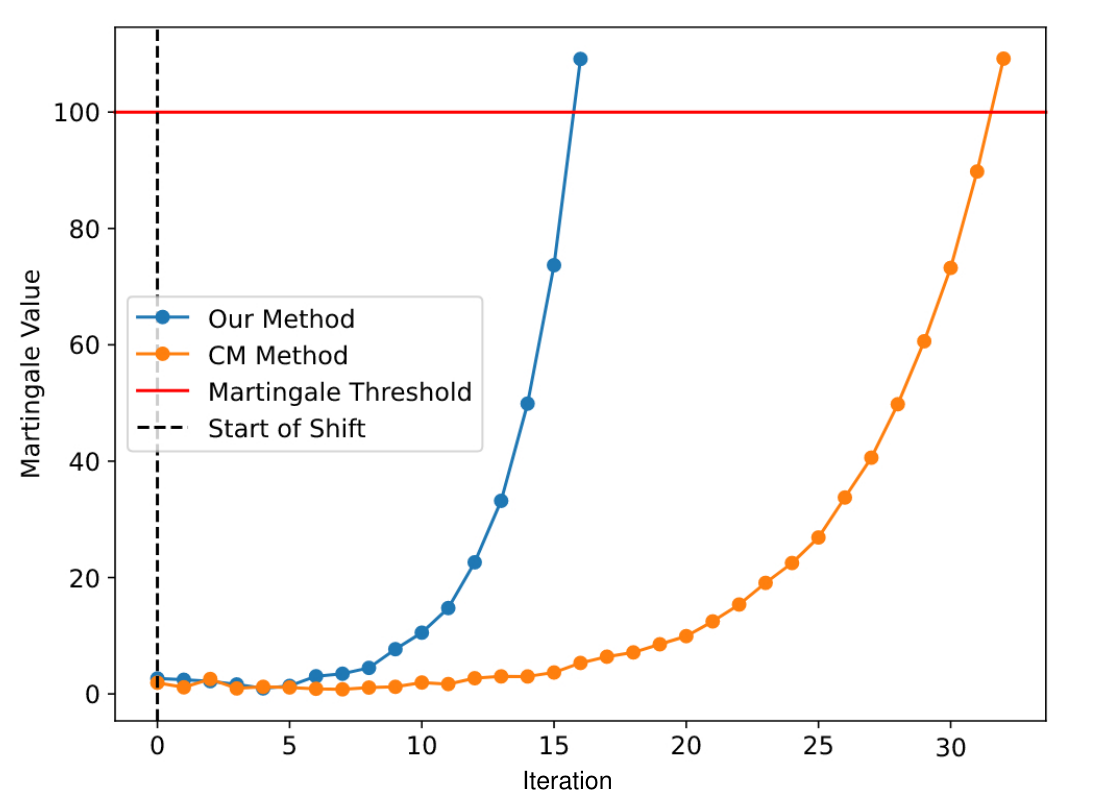}
    \caption{Martingale values for our method and the CM method, in the presence of a sensor degradation shift, which starts occurring at $t=0$. The martingales grow as they detect the shift and an alert is issued when the martingale value exceeds the threshold of 100. Our method raises an alert much sooner on average (14.92 iterations) compared to the CM method (38.10 iterations), showing that our method detects this distribution shift faster than existing methods.}
    \label{fig:martingale_exp1}
\end{figure}
\subsection*{Experiment 1: Fast categorization of distribution shifts}
We first demonstrate that our method is able to detect different types of distribution shifts faster than prior methods. To evaluate our method, we use data from the photorealistic X-Plane 11 flight simulator where an autonomous aircraft uses a PID controller to taxi along the centerline of the runway.

Next, we introduce two types of distribution shifts to the data: first, we introduce Gaussian noise to the images to simulate sensor degradation, and second, we add 200 new sequences of images of the plane taxiing at a new airport, KJFK, that was not in the training dataset (see example images in Figure~\ref{fig:xplane_shifts}). During deployment, images are drawn from either the original sequences, the sequences with added noise, or the sequences from a new runway. To detect whether a shift has occurred and categorize the type of shift, we deploy two stochastic martingales simultaneously to act as runtime monitors. An alert is raised when the martingales reach a threshold of 100. If no alert is raised after 200 samples, the experiment is terminated. The experiment is repeated 100 times.

\begin{table}
    \centering
    \begin{tabularx}{0.5\textwidth}{l *{2}{Y}}
    \toprule
    \multicolumn{3}{c}{\textbf{Mean Iterations Until Alert}} \\
    \multicolumn{1}{l}{Distribution shift}  & Ours & CM \\
    \midrule
    \multicolumn{1}{l}{Sensor degradation}  & \textbf{14.92} &  38.10 \\
    \multicolumn{1}{l}{New environment}   &  \textbf{21.73} & 62.80 \\
    \multicolumn{1}{l}{No shift}  &  none & none \\
    \bottomrule
    \end{tabularx}
    \vspace{3mm}
    \captionof{table}{Experiment 1. Detecting different types of distribution shifts on the X-Plane dataset using our method and the CM method. For each type of distribution shift, we record the average number of iterations before an alert is issued, averaged over 100 trials. Our method is able to detect each distribution shift faster than the CM method.}
    \label{tab:exp1}
\end{table}

We compare our method against the method described by Vovk in \citep{vovk2020testing}, which we will refer to as the conformal martingale (CM) method. For the CM method, we use the nearest distance nonconformity score and the ratio nonconformity score respectively for the two martingales. For each type of distribution shift, we record the number of iterations before an alert is issued by our method and the CM method. The results are summarized in Table~\ref{tab:exp1}. A representative plot showing martingale growth for our method and the CM method is shown in Figure~\ref{fig:martingale_exp1}.

In every scenario, both methods are able to detect the correct distribution shift when one exists (sensor degradation, new environment) and do not issue false alerts when there is no distribution shift, i.e., we empirically observe no false negatives or false positives. Additionally, in every scenario, our method issues an alert corresponding to the correct shift in a fewer number of iterations compared to the CM method, showing that our method is able to detect and categorize distribution shifts more quickly than existing methods.

\subsection*{Experiment 2: Impact of targeted interventions}
Next, we look at how the knowledge of the type of distribution shift enables us to choose the right intervention, which leads to better performance compared to a generic intervention. To evaluate the effect of interventions, we use the same X-Plane dataset as the previous experiment. Images are sampled from the original sequences, the sequences with sensor degradation, or the sequences from a new runway (refer to example images in Figure~\ref{fig:xplane_shifts}). 

\begin{table*}[hb]
\centering
\caption{Experiment 2 - Results from applying different types of interventions on the X-Plane dataset. For each type of distribution shift, we apply either the correct intervention (replacing the sensor if sensor degradation is detected or weighted retraining if an environment shift is detected), the wrong intervention (the same interventions but applied to the converse type of shift), a generic retraining, or no intervention.} 
\begin{tabularx}{\textwidth}{lZ| *{4}{Y}}
\toprule
& & \multicolumn{4}{c}{\textbf{Type of Intervention}}  \\
\multicolumn{1}{l}{Distribution shift}  & Before shift & None & Generic retraining & Wrong intervention  & Correct intervention \\
\midrule
\multicolumn{1}{l}{Sensor degradation}  & 0.17&0.885&0.549&0.428&\textbf{0.175} \\
\multicolumn{1}{l}{New environment}    & 0.17&0.771&0.430&0.759&\textbf{0.225} \\
\bottomrule
\end{tabularx}
\label{tab:exp2}
\end{table*}
If an alert is raised, i.e., a distribution shift is detected by one of our martingales, an intervention is applied. If the alert indicates that sensor degradation has occurred, the correct targeted intervention is to replace the sensor. On the other hand, if the alert indicates that the environment has shifted, the correct targeted intervention is to perform weighted retraining to prioritize learning the most recent samples. We also evaluate a generic intervention, which is retraining the model with equal weight for all samples. After the intervention, the experiment is continued for 100 more samples. Mean squared error (MSE) of the predicted cross-track and heading distance is recorded before the distribution shift and after the intervention. We also record the MSE if no intervention is applied. The comparison of MSE for different types of interventions (over 100 trials) is shown in Table~\ref{tab:exp2}.

Unsurprisingly, the model predictions degrade after a distribution shift occurs, so if no intervention is applied, the MSE increases significantly (from 0.17 to 0.885 or 0.771, depending on the shift). A generic retraining of the model reduces the MSE somewhat. However, the greatest reduction in MSE is achieved if the type of shift is identified and the correct corresponding intervention is applied. 

\subsection*{Experiment 3: System lifecycle performance compared to a scheduled maintenance approach}
Next, we show that applying the correct targeted intervention for each type of distribution shift not only improves model performance in the short-term, but leads to overall better performance and greater safety over the system lifecycle. We posit that a standard approach to maintain lifelong performance, in the absence of runtime monitors for different types of distribution shifts, might be to adopt a standard maintenance schedule, where all interventions are scheduled at a fixed cadence, e.g., every $\gamma$ iterations. We compare such a maintenance schedule against our method, where only targeted interventions are applied, and only when the corresponding shift has been detected.

\begin{table*}[htb]
\centering
\caption{Experiment 3 - Comparison between our method of targeted intervention based on runtime monitoring and a scheduled maintenance approach that applies all interventions at a cadence of $\gamma$ iterations. The model MSE and number of resulting crashes are evaluated over a simulated X-Plane lifecycle, where shifts are induced according to a Poisson process with expected rate of occurrence of a shift $\lambda$. }
\begin{tabularx}{1.0\textwidth}{cc|YYY|YY}
\toprule
& & \multicolumn{3}{c|}{\textbf{MSE}} &  \multicolumn{2}{c}{\textbf{Number of crashes}} \\
$\lambda$ & $\gamma$ & Pre-shift & Ours & Maintenance & Ours & Maintenance\\
\midrule
100                           & 100                          & 0.198                            & \textbf{0.259}                              & 0.301                                     & \textbf{0}                                            & 5.6                                                 \\
100                           & 50                           & 0.198                            & 0.281                              & \textbf{0.248}                                     & \textbf{0}                                            & 3.8                                                 \\
50                            & 50                           & 0.198                            & \textbf{0.377}                              & 0.378                                     & \textbf{0}                                            & 3.5                                                 \\
\bottomrule
\end{tabularx}
\label{tab:exp3}
\end{table*}

We see that our method of only applying targeted interventions when a distribution shift is detected often achieves a lower MSE over the lifecycle as compared to the maintenance schedule, despite the maintenance schedule running \textit{every} intervention at a fixed cadence. When the maintenance frequency ($\gamma$) is equal to the rate at which shifts occur ($\lambda$), our method achieves a lower MSE. Only when maintenance is carried out at double the frequency, i.e., $\gamma$ is half of $\lambda$, does the maintenance schedule outperform our method in terms of lower MSE. 

\begin{figure}
    \centering
    \includegraphics[width=0.5\textwidth]{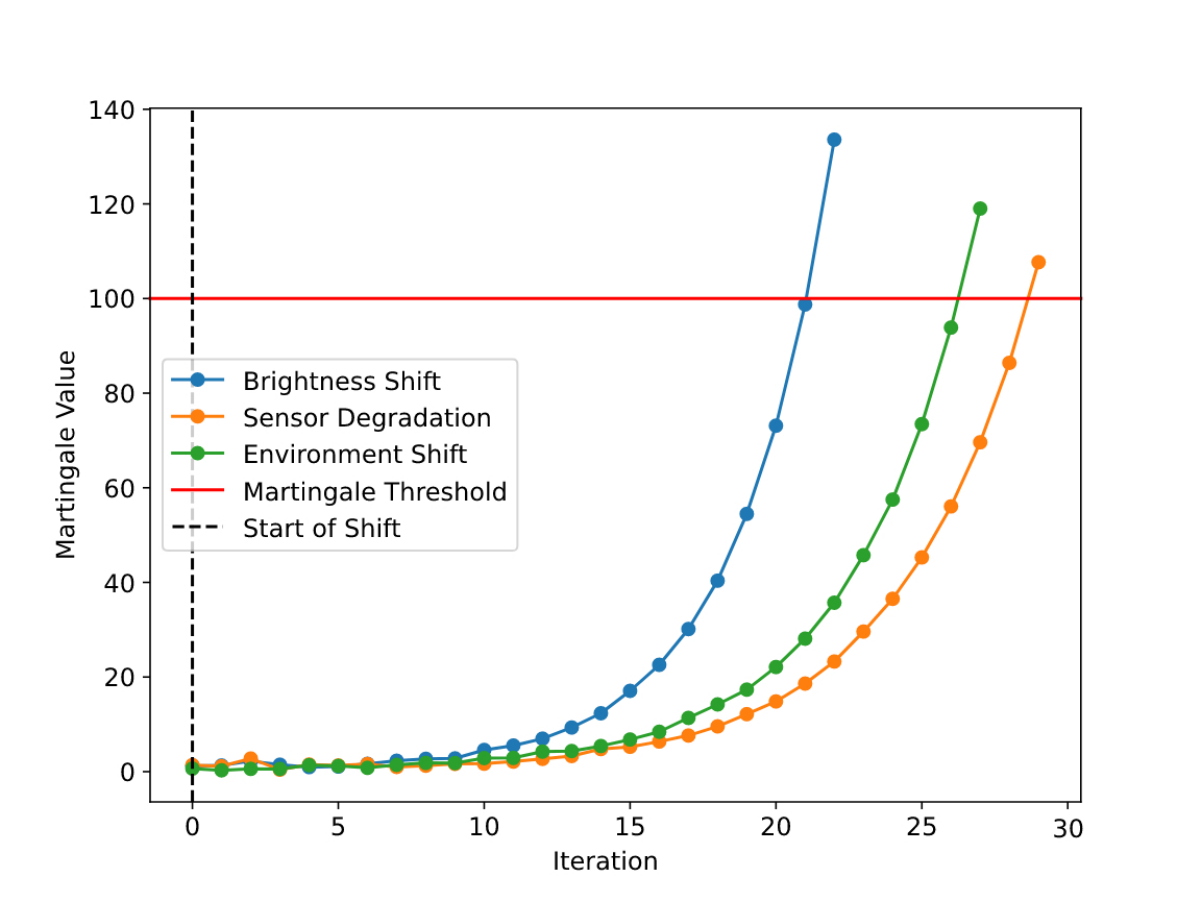}
    \caption{Martingale values for a brightness shift when multiple martingales are deployed simultaneously. All three martingales detect a shift, but the brightness shift martingale issues a warning signal before the other two martingales. }
    \label{fig:martingale_exp4}
\end{figure}

More importantly, our method reduces the frequency of crashes (in fact, completely prevents crashes in this experiment) in all cases while the maintenance schedule occasionally leads to crashes, i.e., on average 3-5 crashes over 1000 iterations. This is because our method uses runtime monitors to quickly detect that a distribution shift has occurred and applies the correct intervention to prevent excessive degradation of the predictions. On the other hand, maintenance is carried out at a fixed cadence, so in some instances where a shift occurs soon after a scheduled maintenance has been completed, that shift can quickly cause errors to snowball and lead to a crash within 45 iterations, before the next scheduled maintenance can take place. Thus, we see that timely and targeted runtime monitoring leads to better performance and fewer failures over the system lifecycle.

\subsection*{Experiment 4: Tailoring custom martingales to specific types of shifts}

Custom martingales can be designed to detect other specific types of shifts. As an example, we define a new martingale to detect a shift in \textit{brightness} of the images, since a brightness shift is a common trigger for existing runtime monitors. This martingale uses the same CNN classifier as before, but it is trained by using the feature vector from the layer of a ResNet-50 architecture that most closely corresponds to image brightness. Specifically, we evaluate all 50 feature vectors from the intermediate layers of the ResNet-50 model and choose the feature vector that changes the most with changes in image brightness. This experiment reinforces one of the advantages of our method: system designers with insights into a likely type of distribution shift can construct a monitor to alert an issue if such a shift occurs.  

\begin{table*}[htbp]
\centering
\caption{Experiment 4 - Comparison of the mean number of iterations before an alert is issued by each of the martingales after each type of distribution shift (sensor degradation - SD, environment shift - ES), and the accuracy with which each distribution shift is identified most quickly by the corresponding martingale.}

\begin{tabularx}{\textwidth}{l |Y Y Y| c}
\toprule
& \multicolumn{3}{c|}{\textbf{Mean Iterations Until Alert}} &  \textbf{First Detection Accuracy} \\
Distribution shift & SD martingale & ES martingale & Brightness martingale & \\
\midrule
Sensor degradation & \textbf{15.51}    & 26.15   & None & 100\%\\
Environment shift & 22.72     & \textbf{16.41}   & None   & 100\%\\
Brightness shift & 22.98    & 28.75    & \textbf{17.72}   & 89\% \\
\bottomrule
\end{tabularx}
\label{tab:exp4}
\end{table*}

On average, all three martingales usually detect their corresponding distribution shift faster than the other two martingales. It is worth noting that a change in brightness also induces a shift in image distribution. Therefore, the brightness martingale detects the shift first 89\% of the time, taking, on average, 17.92 iterations to issue an alert. The other 11\% of time, however, the sensor degradation martingale detects a shift first, because the input distribution is changing as well. Similarly, a shift such as sensor degradation not only changes the distribution of input images but also that of labels, so eventually an alert is raised by the environment shift martingale as well.  Most notably, the brightness martingale only issues an alert for a shift in brightness and not for the other two shifts. This demonstrates that it is possible to design custom martingales that act as runtime monitors for specific predefined distribution shifts, allowing system designers to use their domain expertise to design the most informative runtime monitors.

\subsection*{Experiment 5: Categorizing distribution shifts on a hardware free-flyer space robotics testbed}

\begin{figure}
    \centering
    \begin{minipage}[b]{0.3\textwidth}
        \centering
        \includegraphics[width=\textwidth]{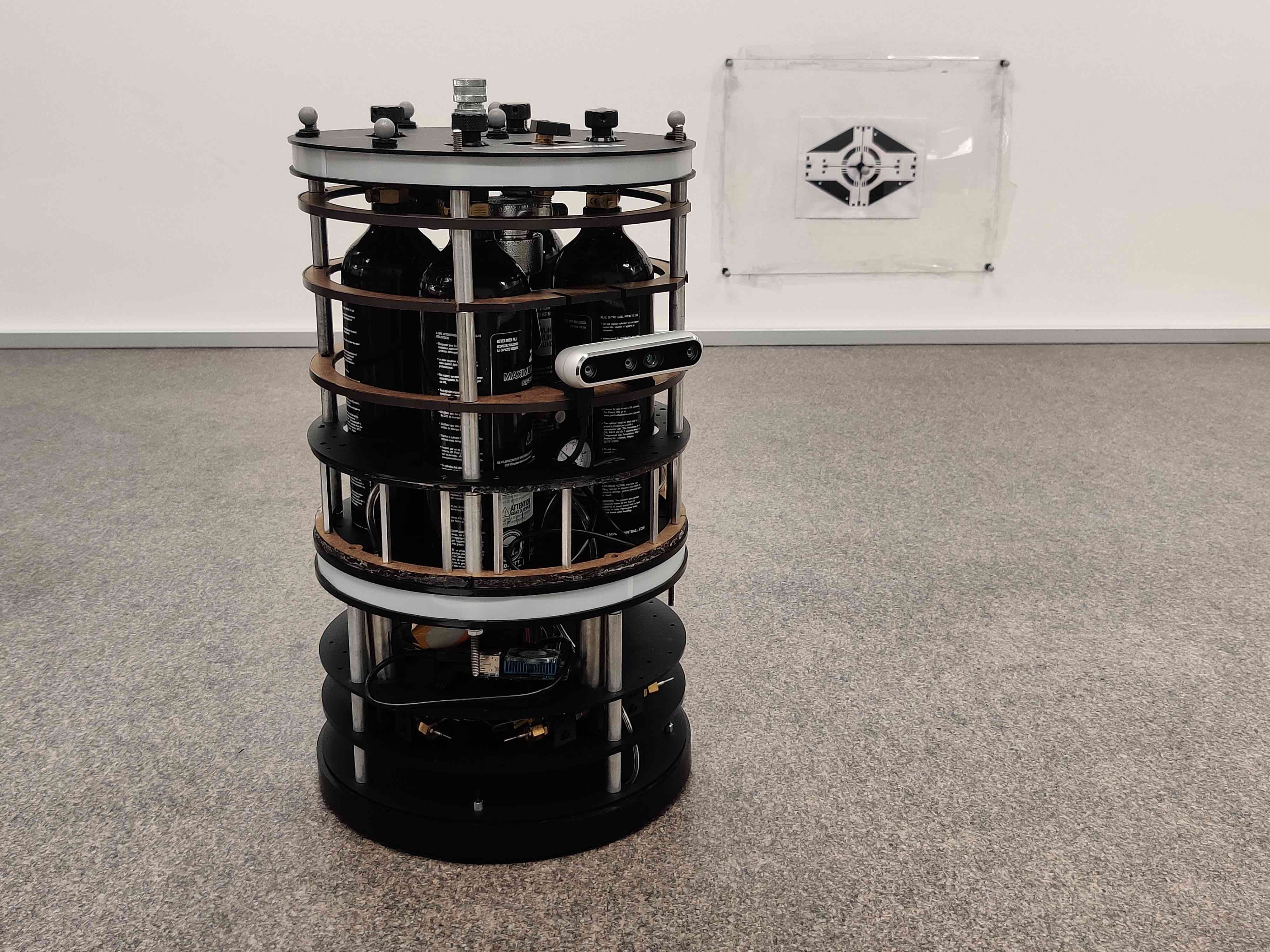}
        \caption{Free-flyer robot platform}
        \label{fig:freeflyer_setup}
    \end{minipage}
    \hspace{0.05\textwidth}
    \begin{minipage}[b]{0.3\textwidth}
        \centering
        \includegraphics[width=\textwidth]{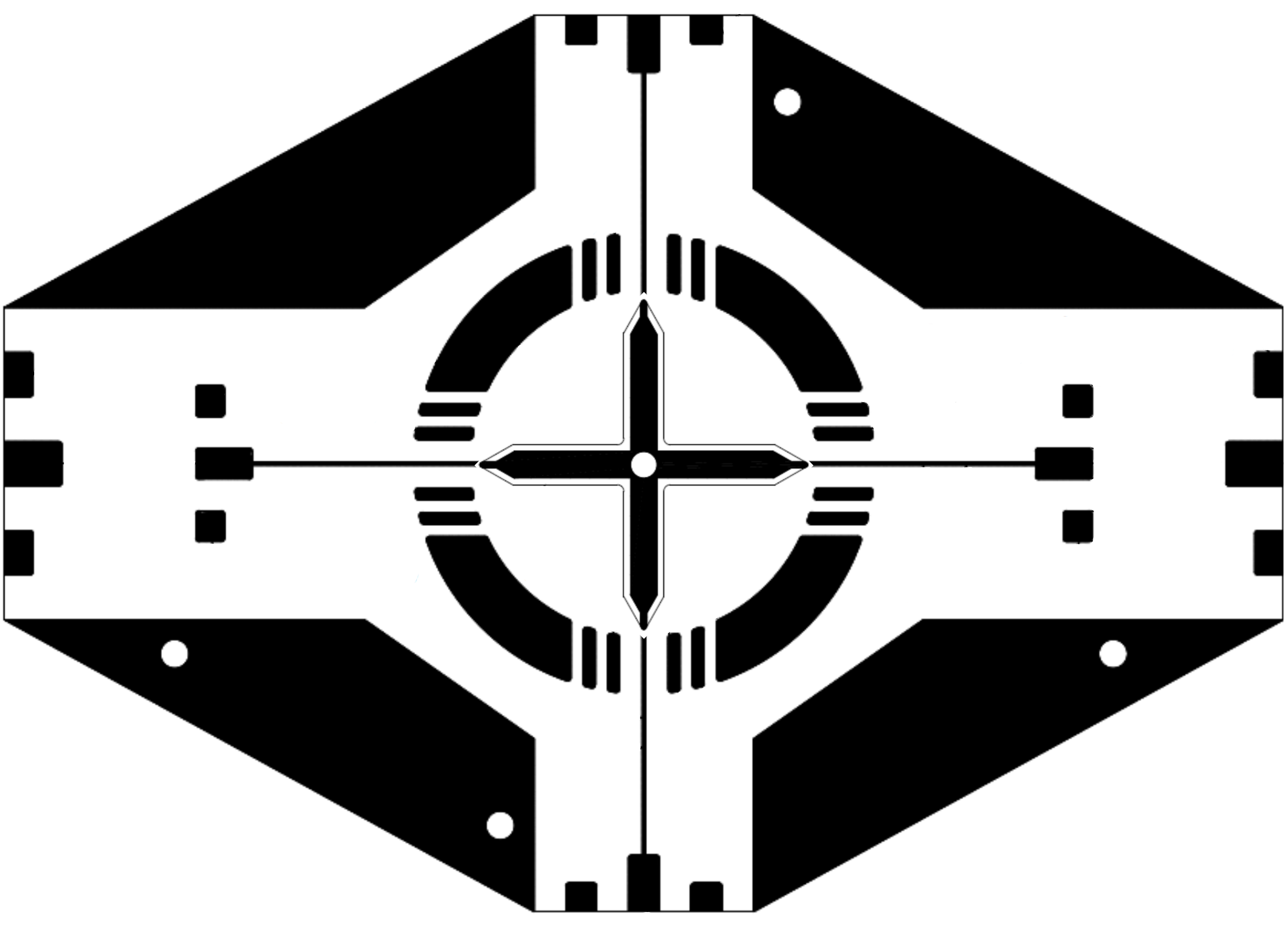}
        \caption{Visual servoing target}
        \label{fig:freeflyer_target}
    \end{minipage}
    \caption{(\ref{fig:freeflyer_setup}) Hardware setup with camera mounted on the side of the mobile robot. (\ref{fig:freeflyer_target}) The visual servoing target that the robot navigates to.}
    \label{fig:freeflyer_platform}
\end{figure}

We also demonstrate the effectiveness of our method on a hardware free-flyer space robotics testbed with input from a forward-facing Intel Realsense D455 camera mounted on the side, as pictured in Figure~\ref{fig:freeflyer_setup}. The free-flyer is a cold gas thruster-actuated 2D mobile robot that floats almost frictionlessly on a smooth granite table, developed to simulate zero-g or zero-friction conditions in aerospace robotics applications \citep{tanaka2020gecko}.

\begin{table}
    \centering
    \begin{tabularx}{0.5\textwidth}{l *{2}{Y}}
    \toprule
    \multicolumn{3}{c}{\textbf{Mean Iterations Until Alert}}  \\
    \multicolumn{1}{l}{Distribution shift}  & Ours & CM \\
    \midrule
    \multicolumn{1}{l}{Sensor degradation}  & \textbf{23.45} &  31.9 \\
    \multicolumn{1}{l}{Environment shift}   &  \textbf{18.20} &  58.3 \\
    \multicolumn{1}{l}{No shift}  &  none & none \\
    \bottomrule
    \end{tabularx}
    \vspace{3mm}
    \caption{Experiment 5 - Detecting different types of distribution shifts on the free-flyer testbed using our method and the CM method. Our method is able to detect each distribution shift faster than the CM method.}
    \label{tab:exp5}
\end{table}

We introduce distributions shifts similar to the previous experiments. We simulate sensor degradation by adding Gaussian noise to the images, an example of which is shown in Figure~\ref{fig:cvr_ff}. We simulate environment shift by changing the starting position (and therefore, the resulting trajectory) of the free-flyer, an example of which is shown in Figure~\ref{fig:labl_ff}. We measure how many time-steps it takes to detect a distribution shift, comparing both our method and the CM method using nearest neightbor nonconformity scores \citep{vovk2020testing}. The results are summarized in Table~\ref{tab:exp5}.

For both the sensor degradation and environment shift, our method detects a shift faster than the CM method. Additionally, neither method issues a false positive, i.e., neither raises an alarm when a shift has not occurred. These results show that our method more rapidly warns us of different types of distribution shifts compared to existing methods and can be used for runtime monitoring in safety-critical contexts.

\section{Conclusion}
\label{sec:conclusion}

In this paper, we introduce a novel framework for diagnosing the cause of a distribution shift, and experimentally validate the speed and adaptability of our method to different types of distribution shifts, models, and datasets. We demonstrate that knowledge of the underlying cause of a distribution shift allows the system designer to induce an appropriate mitigation strategy, and applying our method outperforms a traditional maintenance schedule while reducing cost.

In future work, we would like to explore combining our method with other strategies to prevent performance degradation in real-time systems \citep{SinhaSharmaEtAl2022, sinha2023closing}, and understanding how to combine possibly correlated monitors to achieve stronger system performance guarantees.


\bibliographystyle{isrr2024_format/styles/bibtex/spbasic}
\bibliography{references}

\newpage
\selectfont
\onecolumn
\section*{Appendix}

\section{Overview: Types of Distribution Shift}
    In this section, we clarify the context of our distribution shifts with respect to the well-defined mathematical notions of covariate, concept, and label shift. In a supervised machine learning setting, if $X$ is the input and $Y$ is the output, the training data for a model is a set of samples from the distribution $P(X,Y)$, and the model is learning to approximate the distribution $P(Y|X)$ \cite{10.5555/1462129}. By the definition of conditional probability,
    \begin{align*}
        P(X,Y)  = P(Y|X)P(X) \\ 
                = P(X|Y)P(Y)
    \end{align*} 
    In a \textbf{covariate shift}, $P(X)$ changes, but $P(Y|X)$ remains the same, In a \textbf{label shift}, $P(Y)$ changes, but $P(X|Y)$ remains the same. In a \textbf{concept shift}, $P(Y|X)$, but $P(X)$ remains the same \cite{10.5555/1462129}. However, in practice, multiple shifts may occur jointly, as a change in the $P(X)$ distribution can affect the $P(Y)$ distribution, and vice versa. In this paper, the distribution shifts that were induced do not necessarily fall cleanly into one of these mathematical categories. For example, the sensor degradation shift in Section \ref{sec:experiments} has a change in the $P(X)$ distribution (i.e. a change in the distribution of input images) which our method detects, but no additional information about the $P(Y)$ distribution or $P(Y|X)$ distribution. The new environment shift has a change in $P(Y)$ (i.e. a change in the distribution of trajectory labels) which our method detects, but no additional information about the $P(X)$ distribution or $P(X|Y)$ distribution.

    Note that the nature of the distribution shift detection problem differs from that of the anomaly detection problem, as distribution shift detection focuses on identifying a change in the \textit{distribution} of samples (and it is arguably impossible to make a distributional claim without multiple samples of evidence), whereas anomaly detection focuses on identifying a \textit{single} unusual sample or rare event. Moreover, the intervention method for these two tasks differs significantly. A safety intervention or fallback should be triggered immediately (i.e. within an episode) in response to a sudden change for an unsafe anomalous event. However, interventions for distribution shifts can occur on a somewhat longer timescale (e.g. model retraining), and an alert about distributional shift should inform decision-making about successive deployments.

    The main advantage of our method is as follows: Based on which monitor signaled the distribution shift first, system designers can focus their efforts on improving specific components of the system or gathering additional data. For instance, if an issue is caused by sensor degradation, specific monitors will issue an alert and a good mitigation strategy would be to replace the sensors; if the issue stems from encountering a new environment, other monitors will issue an alert and a good mitigation strategy would be to gather additional data and perform a weighted retraining. In theory, a system designer could apply every possible mitigation strategy each time any distribution shift is encountered, but in practice doing so is costly and inefficient. Ideally, we want to find the best intervention strategy as quickly as possible, which requires that we are able to identify each particular type of distribution shift.
    
\section{X-Plane Experiment Details }
\label{sec:experiments_synthetic}

In this section, we provide further details on the X-Plane experimental setup, including the model architecture that was used and the methodology with which the distribution shifts were induced. 

\subsection*{Model Architecture}
\begin{figure}[h]
    \centering
    \includegraphics[height=0.5\textwidth]{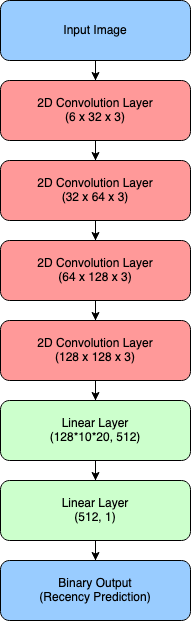}
    \caption{Model architecture used for the detection method during the X-Plane experiments. We use four 2D convolutional layers, followed by two linear layers to produce a binary output indicating which image, cross-track and heading error, or feature vector came first.}
    \label{fig:model-arc}
\end{figure}

For all X-Plane experiments, we use the model architecture outlined in Fig \ref{fig:model-arc} with the Adam Optimizer, a batch size of 32, a fixed learning rate of 1e-4, and a binary cross-entropy loss function. As mentioned in the paper, our method is agnostic to the model used for inference, and we used a simple model to demonstrate the efficacy of our method on small models. For each type of distribution shift, we trained a model with essentially this architecture, except that the input to the model is the cross-track and heading error in the environment shift case and a feature vector of size 100,352 (2048*7*7) in the brightness shift case (this feature vector is the flattened output of one of the intermediate convolutional layers of a ResNet-50 model). The code for running these simulations will be released upon acceptance of this paper.   

\subsection*{Experiment Details: Sensor Degradation}
Using NASA's XPlaneConnect Python API, 1000 simulated video sequences are generated, each corresponding to a taxiing episode of the aircraft on the runway at KSVC airport \citep{xplanedataset}. Each taxiing sequence consists of approximately 30 images of size 200x360x3. These sequences occur at randomly initialized times between 8:00am and 10:00am. We train a classifier to distinguish the ``more recent'' image on these image sequences, by randomly sampling one image from each sequence. The distance to the centerline is estimated by a pre-trained DNN, using images from an outboard camera mounted on the plane. 

\begin{figure*}[h]
    \centering
    \begin{subfigure}[b]{0.325\textwidth}
        \centering
        \includegraphics[width=\textwidth]{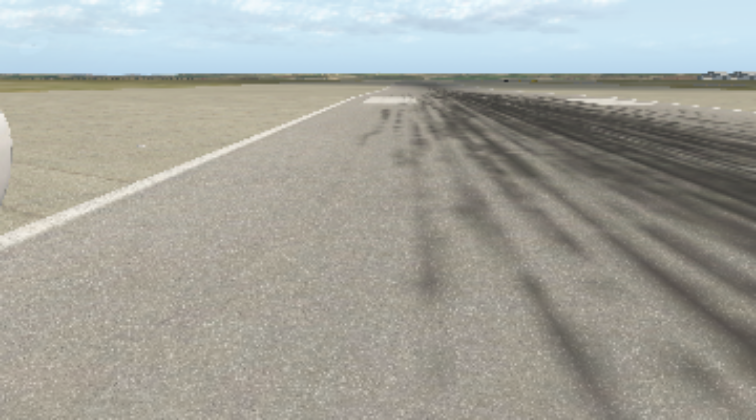}
        \caption{Nominal Image with No Shift}
        \label{fig:nominal}
    \end{subfigure}
    \hfill
    \begin{subfigure}[b]{0.325\textwidth}
        \centering
        \includegraphics[width=\textwidth]{figures/output_image_blurred2.jpg}
        \caption{$\sigma=50$}
        \label{fig:sigma-50}
    \end{subfigure}
    \hfill
    \begin{subfigure}[b]{0.325\textwidth}
        \centering
        \includegraphics[width=\textwidth]{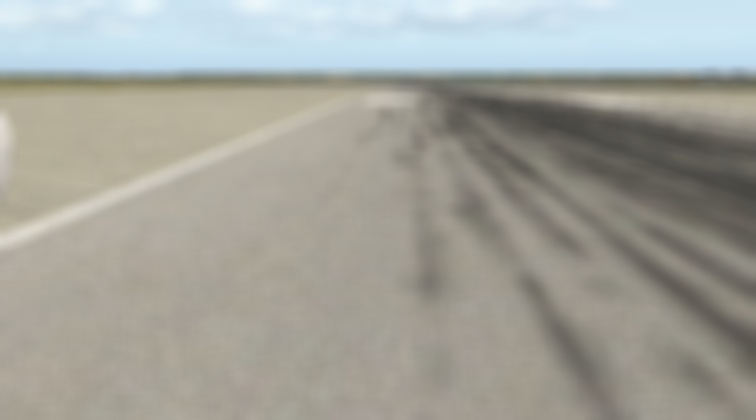}
        \caption{$\sigma=100$}
        \label{fig:sigma-100}
    \end{subfigure}
    \caption{Sample X-Plane 11 images with a distribution shift caused by gradually adding Gaussian noise. Fig \ref{fig:nominal} is the nominal image with no shift, whereas Fig \ref{fig:sigma-50} is the image when $\sigma=50$, and Fig \ref{fig:sigma-100} is the image when $\sigma=100$.} 
    \label{fig:xplane_images}
    \vspace{-3mm}
\end{figure*}

In the sensor degradation experiments, we gradually add Gaussian noise to a set of images in the unseen test set. In the initial experiments, we induce Gaussian noise with a kernel size of (15,15) and a $\sigma$ value that gradually increases. For Experiments 1, 2, 4 and 5, we increase $\sigma$ by 1 at each time step until it reaches the value of $50$. With $\sigma=50$, the Gaussian noise becomes so extreme that our predictor causes the plane to go off the runway after approximately $105$ timesteps. For Experiment 3, we increase $\sigma$ by 2.5 until it reaches a value of $100$. With $\sigma=100$, we observe a crash after approximately $45$ timesteps. We chose a more significant $\sigma$ value in Experiment 3 to simulate a crash if either method does not fix a shift fast enough. When the $\sigma$ value reaches $50$ and $100$ respectively, we continuously sample images blurred with these values. See Fig.~\ref{fig:xplane_images} for example images.

 Note that it takes around 30 minutes to train the initial model on a Macbook Pro M1 chip on the X-Plane dataset of approximately 30,000 images for the sensor degradation shift, and around 2 minutes to retrain after each update step. Each sequence consists of approximately 30 images of size 200x360x3, with 1000 samples observed. 800 of these were used in the training set, while the other 200 were modified according to the Gaussian blur rule defined above.

\subsection*{Experiment Details: Environment Shift}
In the environment shift experiments, we sampled 200 sequences of images from a new runway, KFJK. We induce the same sinusoidal trajectories as before, except with a larger amplitude in order to account for the larger width of the runway. 
In this case, the two linear layers of the neural network had dimension (4, 512) and (512,1), which took approximately 10 minutes to train initially on all 30,000 images, and 45 seconds for each model update. If a warning signal is not issued, the plane will crash after approximately 85 time steps as the predictor will cause the plane to taxi off the runway. 

\subsection*{Experiment Details: Brightness Shift}
In the brightness shift experiments, we sampled 200 sequences with the brightness of the image reduced by 50\%. If a warning signal is not issued, the plane will crash after approximately 120 time steps. The model we used for this method is the same architecture as in Fig. \ref{fig:model-arc}, except that the input to the model is a feature vector from an intermediate layer of a pre-trained ResNet-50 model, rather than the image itself.
 
In order to determine which intermediate layer to use, we constructed a martingale from each intermediate layer of ResNet-50 and ran them simultaneously in order to determine which layer corresponded most closely to image brightness. We experimentally determined that Layer 49 was the fastest in detecting the brightness shift, so we used that martingale when running all of the rest of our experiments. 

\subsection*{Experiment Details: No Shift}
We again take 1000 video sequences of a plane taxiing down a runway using the X-Plane 11 simulator. All sequences occur without a distribution shift, with approximately 30 images of size 200x360x3 in each sequence. We run all of our martingales on these sequences to experimentally show that we issue no false positive warnings when there is no shift.

\subsection*{Experiment Details: System Lifecycle}
We simulate the system lifecycle by constructing sequences of 1000 iterations from the X-Plane dataset and inducing either sensor degradation or environment shift to the data according to a Poisson process with parameter $\lambda$ corresponding to the expected rate of occurrence of the shift. As before, sensor degradation is simulated by adding noise to the images and environment shift is simulated by introducing images from the new runway. The shifts are large enough that if an intervention is not applied within 45 iterations after the shift occurs, the degraded model predictions cause the aircraft to deviate off the runway and crash.

\section{Free-Flyer Hardware Experiments}\label{ap:free-flyer}

\begin{figure*}[ht]
    \centering
    \begin{subfigure}[c]{0.325\textwidth}
        \centering
        \includegraphics[width=\textwidth, height=3.3cm]{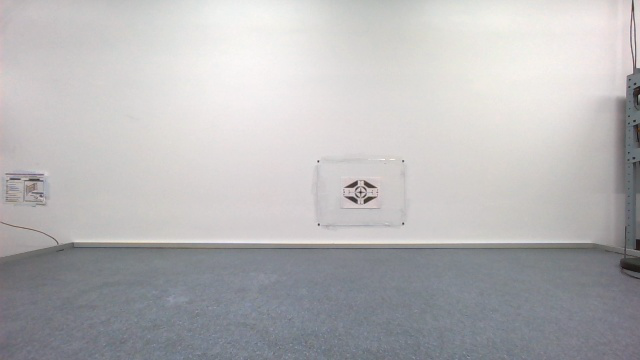}\vspace{1mm}
        \caption{Initial image}
        \label{fig:cam_ff}
    \end{subfigure}
    \begin{subfigure}[c]{0.325\textwidth}
        \centering
        \includegraphics[width=\textwidth, height=3.3cm]{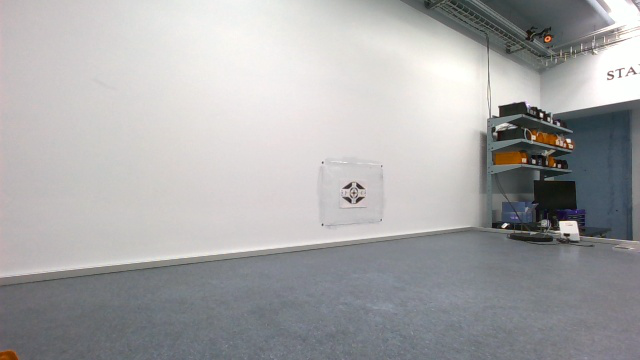}\vspace{1mm}
        \caption{Environment Shift}
        \label{fig:labl_ff}
    \end{subfigure}
    \begin{subfigure}[c]{0.305\textwidth}
        \centering
        \includegraphics[width=\textwidth, height=3.3cm]{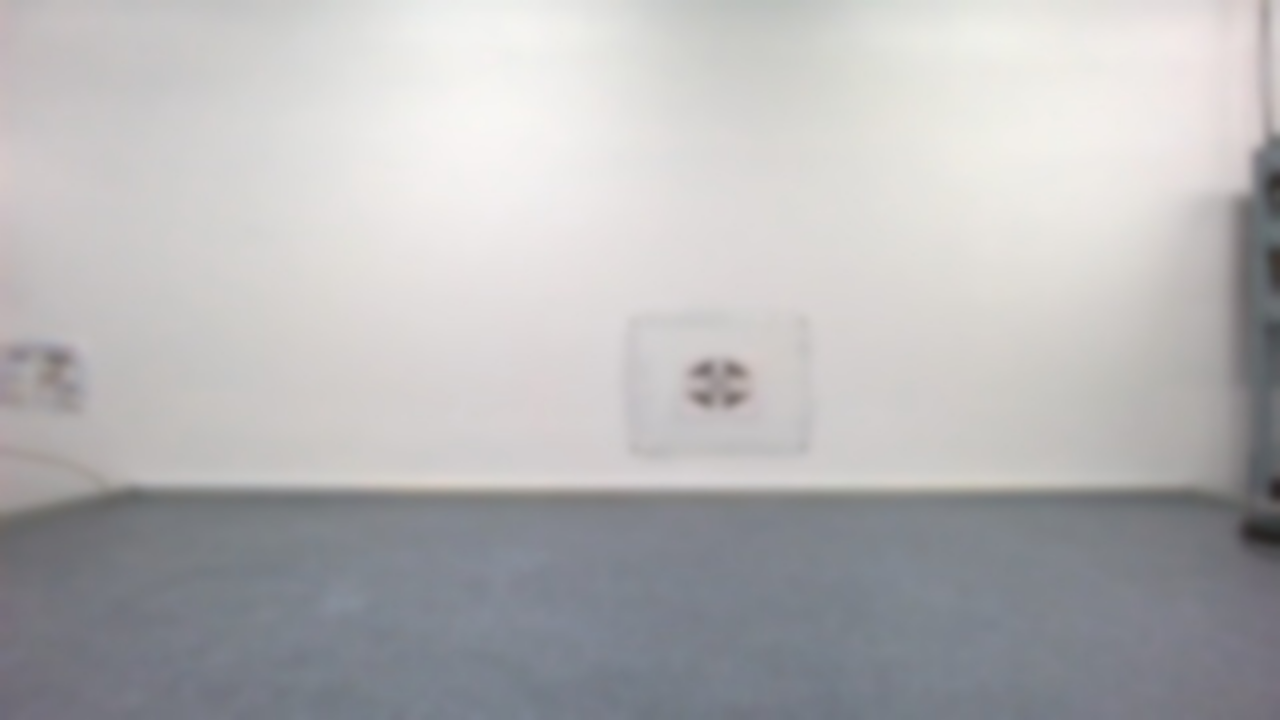}\vspace{1mm}
        \caption{Sensor Degradation}
        \label{fig:cvr_ff}
    \end{subfigure}
    \caption{Images generated from the free-flyer visual servo, with (\ref{fig:cam_ff}) a standard camera angle,  (\ref{fig:labl_ff}) a environment shift, and (\ref{fig:cvr_ff}) a sensor degradation shift. The trajectory of the free-flyer ends at the visual servoing target.   }
    
\label{fig:free-flyer-images}
\end{figure*}

We collect data from multiple episodes of the free-flyer moving from a starting position to the docking target. This data consists of 9000 camera images from 30 episodes (each image is of size 360x640x3) as well as the associated ground truth relative pose of the free-flyer with respect to the docking target at each iterations, which is recorded using a motion capture system above the free-flyer testbed. These images are used to train a classifier to distinguish the ``more recent" image. An example image is shown in Figure~\ref{fig:cam_ff}.

In these experiments, the free-flyer performs a learning-based visual servoing task that emulates autonomous spacecraft docking. Images from the onboard camera are used to guide the free-flyer from a starting point towards a static visual target (as shown in Figure~\ref{fig:freeflyer_target}), which is a pattern defined by the International Docking System Standard for spacecraft docking adapters. 

\begin{wrapfigure}{R}{0.5\textwidth}
    \centering
    \includegraphics[width=0.45\textwidth]{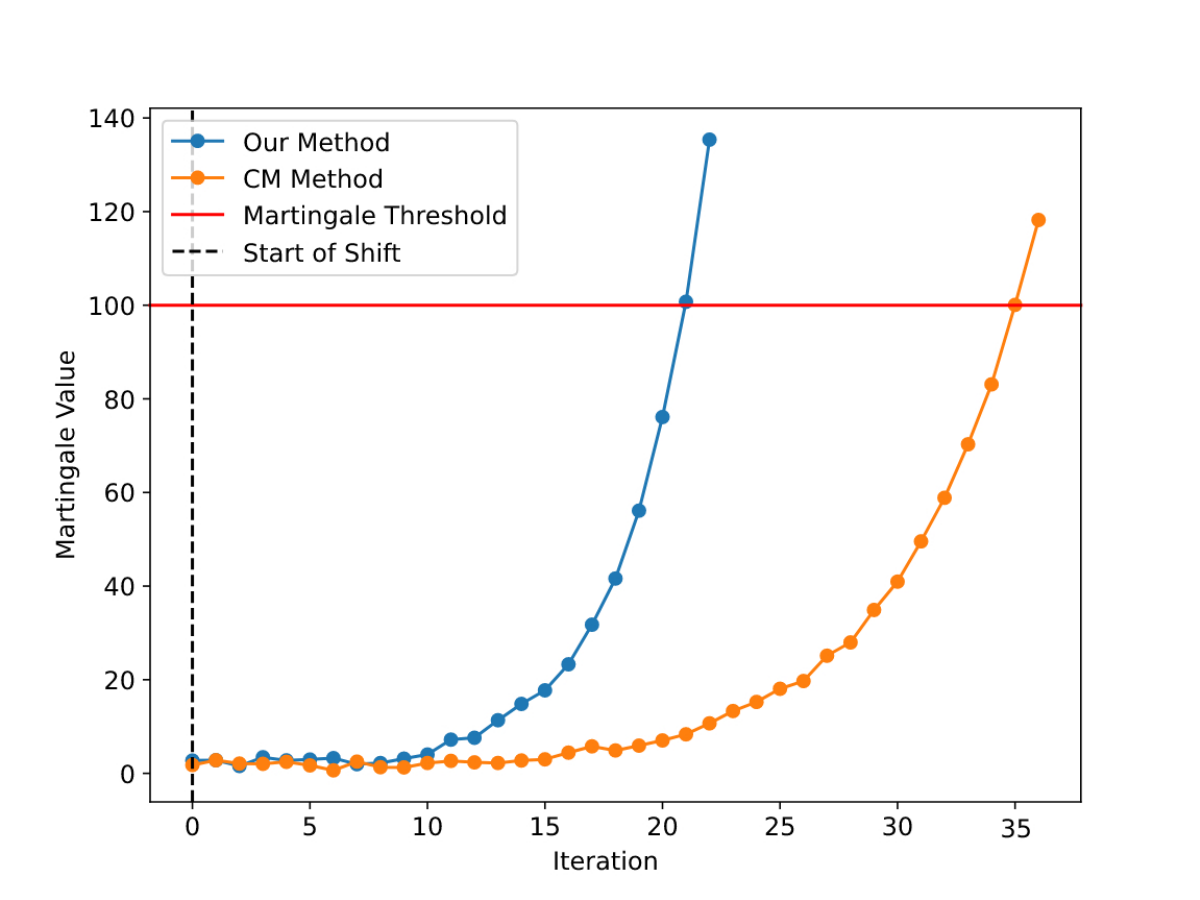}
    \caption{Martingale values growing to indicate a distribution shift, specifically sensor degradation, in a real-time robotics setting recorded on the free-flyer hardware. The martingale associated with our method grows much faster than the martingale associated with the CM method, demonstrating that our method is able to issue an alert faster than existing methods.}
    \label{fig:martingale_exp5}
\end{wrapfigure}

To collect the initial image data, we positioned the robot at or around a fixed starting point on the granite table while ensuring that the visual target stayed within the field of view of the camera. For the new environment distribution shift, we collected data by positioning the robot at a different starting point about one meter away from the original starting point. We captured 30 sequences from each starting point with around 300 images per sequence for a total of 18,000 images, as well as the corresponding $x,y$ position of the freeflyer using the Optitrack motion capture system. For the sensor degradation experiment, we add Gaussian noise to the data from the original starting point using the same process as described above for the X-Plane experiments with $\sigma=50$. For both sets of experiments, we use the model architecture shown in Fig.~\ref{fig:model-arc}, except that instead of predicting the cross-track and heading error in the environment shift case, we predict the $x,y$ position of the free-flyer on the granite table relative to the visual servoing target. A representative plot of the martingale growth for our method and the CM method is shown in Figure~\ref{fig:martingale_exp5}.

\end{document}